\documentclass[a4paper,conference]{IEEEtran}
\usepackage{ amssymb }
\usepackage{graphicx,color,psfrag}
\usepackage{multirow}
\usepackage{booktabs}
\usepackage{amsfonts} 
\usepackage{listings}
\usepackage{paralist}
\usepackage{amsmath}
\usepackage[hyphens]{url}
\usepackage{verbatim}
\usepackage{enumitem}
\usepackage{footnote}
\usepackage{threeparttable}
\usepackage{hyperref}
\usepackage{etoolbox}
\usepackage{multicol}
\usepackage{ tipa }
\usepackage{float}
\usepackage{mathtools}
\usepackage{varioref}
\usepackage{adjustbox}
\usepackage{rotating}
\usepackage{epstopdf}
\usepackage{bm}
\usepackage{caption}
\usepackage{subcaption}
\usepackage{breqn}
\usepackage{subfloat}
\DeclareMathOperator*{\argminA}{arg\,min}

\begin{document}

\title{Efficient training of lightweight neural networks using Online Self-Acquired Knowledge Distillation}
\author{\IEEEauthorblockN{Maria Tzelepi and Anastasios Tefas \\}
\IEEEauthorblockA{Department of Informatics\\
Aristotle University of Thessaloniki\\
Thessaloniki, Greece\\
Email: $\{$mtzelepi,tefas$\}$@csd.auth.gr}}

\maketitle

\begin{abstract}
Knowledge Distillation has been established as a highly promising approach for training compact and faster models by transferring knowledge from heavyweight and powerful models. However, KD in its conventional version constitutes an enduring, computationally and memory demanding process. In this paper, \textit{Online Self-Acquired Knowledge Distillation} (OSAKD) is proposed, aiming to improve the performance of any deep neural model in an online manner. We utilize k-nn non-parametric density estimation technique for estimating the unknown probability distributions of the data samples in the output feature space. This allows us for directly estimating the posterior class probabilities of the data samples, and we use them as soft labels that encode explicit information about the similarities of the data with the classes, negligibly affecting the computational cost. The experimental evaluation on four datasets validates the effectiveness of proposed method.
\end{abstract}

\section{Introduction}
Deep learning models \cite{deng2014tutorial} have been extensively utilized over the recent years so as to resolve a broad range of computer vision tasks \cite{guo2016deep}. However, their superior performance usually stems from their depth and complexity. This constitutes a major impediment on applying these models for real-time applications as well as on devices with limited computational resources such as mobile phones and embedded systems. Thus, an apparent need for developing compact yet effective models has been arisen.
Among other solutions proposed to achieve this goal \cite{cheng2017survey}, \textit{Knowledge Distillation} (KD) \cite{hinton2015distilling} has emerged as a very effective way to address this issue.

KD, in its seminal approach, describes the procedure where the knowledge of a high-capacity model, known as teacher, which exhibits high performance, is transferred to a more compact and faster model, known as student. The student model is trained to regress the so-called \textit{soft labels} produced by the teacher by raising the temperature of the softmax activation function on the output layer of the network. These soft labels convey more information of the way the model learns to generalize, as compared to the hard labels, trying implicitly to reveal similarities over the data. Other approaches include the transfer of knowledge from teachers to students of identical capacity \cite{Yim_2017_CVPR,Furlanello2018BornAN,lan2018self}. The latter procedure is known as \textit{self-distillation}.

KD methods can be classified into two broad categories: \textit{online} and \textit{offline} KD. The latter is what has already been described as the multi-stage procedure of first training a more complex and powerful teacher model and then distilling the knowledge to a simpler student model by training it to mimic the teacher. On the contrary, \textit{online} KD describes the procedure where the teacher and the student models are trained simultaneously, that is without the stage of pre-training the teacher network. Online KD includes works proposing to train multiple models mutually from each other \cite{Zhang_2018_CVPR}, as well as works proposing to build ensembles of multiple identical branches of a target model in order to build a strong teacher and distill the knowledge from the teacher to the target model \cite{NIPS2018_7980}. 

In this work, we pursue the direction of online KD through a different perspective: We propose a single-stage self-acquired knowledge distillation approach, namely \textit{Online Self-Acquired Knowledge Distillation} (OSAKD) for improving the performance of any deep neural model in an online manner. The motivation behind this work is to train effectively small networks with an additional supervision that conveys extra knowledge beyond the hard labels from the model itself and also in an online manner. That is, in a typical classification problem the softmax classifier, which produces a probability distribution over the classes, inherently suppresses all the activations excluding the highest one.  Conventional KD methods manifest that the class probability distribution of a strong teacher model provides useful information on the similarities of the data with all the classes. Thus, instead of training a simpler student model merely with the hard labels, it is useful to maintain the similarities of the samples with the classes \cite{hinton2015distilling}. In addition, in \cite{Furlanello2018BornAN} it is shown that useful information about the similarities of the samples with the classes can be obtained even by transferring the knowledge through the class probability distribution from a teacher of identical capacity to student. Furthermore, in \cite{NIPS2014_5484} it is stated that small networks usually have the same representation capacity as large networks, however they are harder to train, compared to large networks. Thus, taking the aforementioned observations into consideration, the question that arises is how can someone efficiently train small yet effective networks, deriving additional information about the similarities between the classes from the model itself and also in an online manner.

Towards this end, we utilize the k-nn non-parametric density estimation technique \cite{duda2012pattern} for estimating the unknown probability distributions of the data samples in the feature space of the output layer. In this way, we are able to directly estimate the posterior class probabilities of the data samples, and use them as soft labels. The estimated soft labels explicitly encode information on the similarity of each training sample with the classes, while it is expected that as the training progresses we learn more and more reliable and meaningful soft labels, since they are driven by the supervised loss. This argument is also experimentally validated. Furthermore, it should be emphasized that as opposed to a general classification task where when estimating the posterior class probabilities some errors lead to the accumulation of errors, and the goal is to minimize them, in our case, propose to estimate the posterior class probabilities utilizing the k-nn density estimation technique, so as to use them as soft labels, that is as an auxiliary task to the conventional supervised loss. Thus, our goal is not to directly minimize the estimation error for them, but rather to continuously estimating and considering them through the training process, so as to assist the main classification objective.

The intuition of the proposed online distillation method, considering a probabilistic view of KD is as follows: Deep neural models transform the probability distribution of the data, layer by layer, learning increasingly complex layer representations. Considering a multi-class classification problem, a conventional supervised loss, forces the data representations in the output layer of a neural model to become one-hot representations. However, trying to convert the complex data representations to one-hot representations usually leads to over-training and also requires deeper and more complicated models. Thus, in this work considering the neighborhood of each sample as it is explained in the subsequent Section, we produce the soft labels that encode the similarity of each sample to each of the classes. That is, the hard label objective forces all the data representations belonging to a certain class to concentrate one point. On the contrary, each data representation is forced to approach its own soft label encoding its similarity with the classes of the problem. In this way, we learn representations that recover the similarities of the samples with the classes, improving the generalization ability of the model.

The main contributions of this paper can be summarized as follows:\\
-We propose a novel \textit{Online Self-Acquired Knowledge Distillation} method.\\
-The OSAKD method acquires the soft labels from the model itself in an online manner, without requiring the utilization of multiple models or copies of the model like the most online KD methods, rendering it more efficient. \\
-The proposed method is model-agnostic, that is, it is applicable to several neural network architectures.\\
-The experimental evaluation indicates the OSAKD method can improve the classification performance of simple and deeper models.\\

The reminder of the manuscript is organized as follows. Section \ref{s2} discusses relevant works. The proposed OSAKD method is presented in detail in Section \ref{s3}. Subsequently, in Section \ref{s4} the experiments conducted to validate the proposed method are provided, and finally the conclusions are drawn in Section \ref{s5}.

%
%
\section{Prior Work}\label{s2}

In this Section we discuss prior studies on self-distillation and online KD. Regarding the first case, a self-distillation approach where a teacher model is initially trained, and then after its convergence, an identical student model is trained with both the goals of the hard labels and matching the output of the teacher model, however without softening the logits (i.e., the inputs to the final softmax activation function) by raising the temperature, is proposed in \cite{Furlanello2018BornAN}. Similarly, a target model is trained with a conventional supervised loss, the self-discovered knowledge is extracted, and in the second training stage, the model is trained with both the supervised and the  distillation losses in \cite{lan2018self}. 

In this paper, a self-distillation method is also proposed. However, a key attribute of the proposed method is that the knowledge is distilled within the same model online. The proposed approach does not use  the aforementioned multiple stages of the training pipeline, which renders it more efficient. It also uses soft labels explicitly, recovering the similarities of each training sample with the classes.

Subsequently, several works proposing online distillation have been also recently proposed. A method namely codistillation has been proposed in \cite{46642}. It improves the accuracy by proposing to train $k$ copies of a target model in parallel, by adding a distillation term to the loss function of the $i$-th model to match the average prediction of the other models. A quite similar approach is proposed in \cite{Zhang_2018_CVPR} where an ensemble of students teach each other throughout the training process. That is, each student is trained with a conventional supervised learning loss, and a distillation loss that aligns each student's class posterior with the class probabilities of other students. In this way, each model acts as a teacher of the other models. In this approach, as opposed to the aforementioned codistillation method \cite{46642}, different networks can be used for the mutual training.

Furthermore, an online distillation approach where a multi-branch version of the network is built by adding identical branches each of which constitutes an independent classification model with shared low level layers, and a strong teacher model is created utilizing a gated logit ensemble of the multiple branches, in \cite{NIPS2018_7980}. Each branch is trained with the conventional classification loss and the distillation loss which regresses the teacher's prediction distributions. Finally, a recent work \cite{kim2019feature}, combines the previous works \cite{Zhang_2018_CVPR} and \cite{NIPS2018_7980} proposing an online mutual knowledge distillation method for enhancing both the performance of the fusion module and the sub-networks. That is, when different sub-networks are used, the sub-networks are trained similar to \cite{Zhang_2018_CVPR}, while when identical sub-networks are used, the low level layers are shared, and a multi-branch architecture similar to \cite{NIPS2018_7980} is used. 

Contrariwise, we propose a self-supervised online KD methodology which allows synchronous model updating, without the need of building multiple identical models, or using multiple possibly different models to mutually teach each other, which comes with additional computational cost.

%
%
\section{Proposed Method}\label{s3}
We consider a $C$-class classification problem, 
and the labeled data ${\{\mathbf{x}_i,\mathbf{y}_i \}}_{i=1}^{N}$, where $\mathbf{x}_i\in\Re^{D}$ an input vector and $D$ its dimensionality, while $\mathbf{y}_i \in  \mathcal{Z^C}$ corresponds to its $C$-dimensional one-hot class label vector. 
For an input space $\mathcal{X} \subseteq \Re^{D}$ and an output space $ \mathcal{F} \subseteq \Re^{C}$, we consider as $\phi(\cdot\, ; \mathcal{W}):\mathcal{X} \rightarrow \mathcal{F}$ a deep neural network with $N_L \in \mathbb{N}$ layers, and set of parameters $\mathcal{W}=\{\mathbf{W}_1, \dots, \mathbf{W}_{N_L}\}$, where $\mathbf{W}_L$ are the weights of a specific layer $L$, which transforms its input vector to a $C$-dimensional probability vector. That is, $\phi(\mathbf{x}_i\, ; \mathcal{W}) \in \mathcal{F}$ corresponds to the output vector of $\mathbf{x}_i \in \mathcal{X}$ given by the network $\phi$ with parameters $\mathcal{W}$.

Thus, considering the typical classification problem, we seek for the parameters $\mathcal{W^*}$ that minimize the cross entropy loss, $\mathcal{L}_{ce}$, between the output vector $\phi(\mathbf{x}_i\, ; \mathcal{W})$ and the one-hot class label vector $\mathbf{y}_i$:\\
\begin{equation}
\mathcal{W^*}= \argminA_{\mathcal{W}} \sum_{i=1}^{N} \mathcal{L}_{ce}\big(\mathbf{y}_i, \phi(\mathbf{x}_i\, ; \mathcal{W})\big),
\end{equation}
The cross entropy loss for a sample $i$ is formulated as:\\
\begin{equation}
\mathcal{L}_{ce}(\mathbf{y}_i,\mathbf{z}_i) = \sum_{m=1}^{C} y_i^m \log(z_i^m  ),
\end{equation}
where $y_i^m$ is the $m$-th element of $\mathbf{y}_i$ one-hot label vector, and
$z_i^m $ is defined as the output of the softmax operation on the $C$-dimensional network's output:\\
\begin{equation}\label{eq_z}
z_i^m  = \frac{\exp({\phi(\mathbf{x}_i\, ; \mathcal{W}})^m)}{\sum_{j=1}^{C}\exp({\phi(\mathbf{x}_i\, ; \mathcal{W})^j)}}.
\end{equation} 
In this work, we propose to distill additional knowledge online from the model itself throughout the network's training. To this end, we propose  to utilize k-nn non-parametric density estimation \cite{duda2012pattern} for estimating the unknown probability distributions of the data samples in the output space $ \mathcal{F} \subseteq \Re^{C}$. Generally, the idea of non-parametric density estimation relies on the fact that the probability $P$ that a vector $\mathbf{x}$ will fall in a region $R$ is given by: 
$P= \int_{R} p(\mathbf{x}')d\mathbf{x}'$.
Hence, $P$ is a smoothed version of the density function $p(\mathbf{x})$ and this smoothed value of $p$ can be estimated by estimating the probability $P$. Considering, that $N$ samples ${\{\mathbf{x}_i}\}_{i=1}^{N}$ are drawn independently and identically distributed according to the probability law $p(\mathbf{x})$, then the probability that $k$ of these $N$ fall in $R$ is given by the binomial law: $P_k= {N\choose k} P^k(1-P)^{N-k}$, and the expected value for $k$ is given by: $\mathbb{E}[k]= NP$.
 
Moreover, this binomial distribution for $k$ peaks very sharply about the mean, so that it is expected that the ratio $\frac{k}{N}$ will be a very good estimate for the probability $P$, and therefore for the smoothed density function. The larger the $N$, the more accurate the estimate is. Assuming also that $p(\mathbf{x})$ is continuous and that the region $R$ is so small that $p$ does not considerably vary within it, we can arrive at the following equation: $\int_{R} p(\mathbf{x}')d\mathbf{x}' \simeq p(\mathbf{x}) V$, 
where $\mathbf{x}$ is a point within $R$ and $V$ is the volume enclosed by $R$. By combining the previous equations we arrive at the following estimate for $p(\mathbf{x})$: $p(\mathbf{x}) = \frac{\frac{k}{N}}{V}$.

In the k-nn density estimation we determine the number of nearest neighbors, $k$, and we adjust the volume, as compared to the parzen windows method where for a fixed-length volume $V$ we  observe how many points $k$ fall into the region.

The key advantage of the k-nn density estimation method is that it allow us to directly estimate the posterior probabilities $P(c_m |\mathbf{x})$ of the class being $c_m, m=\{1, \cdots, C\}$, from a set of $N$ labelled data by using the samples to estimate the densities involved, \cite{duda2012pattern}. That is, assuming that we place a cell of volume $V$ around $\mathbf{x}$ and capture $k$ samples, $k_m$ of which belong to the class $c_m$. Then, the estimate for the probability $p(\mathbf{x}| c_m)$ is: $p(\mathbf{x}| c_m)=\frac{k_m}{N_mV}$,
where $N_m$ is the number of samples which belong to the class $c_m$, and similarly the unconditional density is estimated as: $p(\mathbf{x})=\frac{k}{NV}$, and the priors can be approximated by: $P(c_m)=\frac{N_m}{N}$.

Thus, the posterior probabilities using the Bayes rule are given by:
\begin{equation}
P(c_m|\mathbf{x})= \frac{p(\mathbf{x}| c_m)P(c_m)}{p(\mathbf{x})} = \frac{\frac{k_m}{N_mV}\frac{N_m}{N}}{\frac{k}{NV}}=\frac{k_m}{k}.
\end{equation}

Therefore, for a specific number of nearest neighbors, $k$, the posterior probabilities can be estimated and used as soft labels for the network's training. That is, the soft label for a sample $i$ is formulated as: $\mathbf{s}_i = \big[\frac{k_1}{k}, \frac{k_2}{k}, \dots, \frac{k_C}{k}\big].$

Thus, in the proposed distillation training procedure, we seek for the parameters $\mathcal{W^*}$ that minimize the overall loss of cross entropy, $\mathcal{L}_{ce}$ and self-distillation, $\mathcal{L}_{sd}$:\\
\small 
\begin{equation}\label{eql}
\mathcal{W^*}= \argminA_{\mathcal{W}} \sum_{i=1}^{N} [\lambda \mathcal{L}_{ce}\big(\mathbf{y}_i, \phi(\mathbf{x}_i\, ; \mathcal{W})\big) + (1-\lambda) \mathcal{L}_{sd}\big(\mathbf{s}_i, \phi(\mathbf{x}_i\, ; \mathcal{W})\big)],
\end{equation}
\normalsize
where $\lambda$ balances the importance between predicting the hard labels in $\mathbf{y}_i$ and regressing the soft labels in $\mathbf{s}_i$.  Either Kullback-Leibler (KL) divergence or the Mean Squared Error (MSE) between the soft labels, and the actual predictive vectors can be utilized to train the network. In this work, we utilize MSE.

Consequently, the OSAKD training procedure is as follows. The input images are fed to the network, and for each sample the class predictions are produced. Subsequently, the soft labels are computed based on the neighborhood of each sample, according to the procedure described above. Then, the network is trained using the cross entropy loss with the hard labels, and concurrently using the distillation loss so as to regress the produced soft labels, enforcing it to regard the similarity of each sample with the classes.


%
%
\section{Experiments}\label{s4}
Four datasets were used to validate the performance of the proposed method. That is, Cifar-10 \cite{krizhevsky2009learning}, Street View House Numbers (SVHN) \cite{netzer2011reading}, Fashion MNIST \cite{xiao2017fashionmnist}, and Tiny ImageNet \footnote{https://tiny-imagenet.herokuapp.com/}. The descriptions of the models' architectures used for each dataset follow bellow. Test accuracy was used to evaluate the proposed method. Each experiment is repeated five times and the mean value and the standard deviation are reported, considering the maximum value of test accuracy for each experiment. The curves of mean test accuracy are also provided. Finally, the sum of floating point operations (FLOPs) is used to evaluate the complexity of the proposed method.

\subsection{CNN Models}
In the case of the Cifar-10, and SVHN datasets a simple CNN model is utilized, consisting of five layers; two convolutional layers with 6 filters of size $5 \times 5$ and 16 filters of size $5 \times 5$ respectively, followed by a Rectified Linear Unit (ReLU) \cite{nair2010rectified} activation, and three fully connected layers ($128 \times 64 \times 10$).  The convolutional layers are followed by a $2 \times 2$ max-pooling layer with a stride of 2. In the first two fully connected layers the activation function is the ReLU, while the last output layer is a 10-way softmax layer which produces a distribution over the 10 class labels of the utilized datasets. In the case of the Fashion MNIST dataset  a simple architecture is also utilized, consisting of two convolutional with 20 filters of size $5 \times 5$ and 50 filters of size $5 \times 5$ followed by a ReLU activation, and two fully connected layers ($64 \times 10$). The convolutional layers are followed by  $2 \times 2$ max-pooling layer with a stride of 2. In the fist fully connected layer a ReLU activation is applied, while the last output layer is a 10-way softmax layer. 
Finally, in the challenging case of Tiny ImageNet dataset the common ResNet-50 \cite{DBLP:journals/corr/HeZRS15} architecture is utilized, without utilizing any pre-trained model in order to avoid image resizing. It is noteworthy that the goal of this work is not to provide state-of-the-art models, but rather to  to evaluate the effect of the proposed online self-distillation method on training lightweight model that can be effectively deployed on embedded and mobile devices. To this end, we use the aforementioned simple CNN architectures in three out of four cases, while we also utilize a common powerful network in the case of Tiny ImageNet, validating our claim that the proposed method is model agnostic. Finally, for comparison purposes against previous online KD approaches, we also utilize ResNet-32 \cite{DBLP:journals/corr/HeZRS15} to perform experiments on Cifar-10 dataset.

\subsection{Implementation Details} 
The mini-batch gradient descent is used for the networks' training, considering mini-batch size of 64 samples. The learning rate is set to $10^{-3}$, and the momentum is 0.9. The models are trained on an NVIDIA GeForce GTX 1080 with 8GB of GPU memory for 100 epochs. The parameter $\lambda$ in eq. (\ref{eql}) for controlling the balance between the contributing losses, is set to 0.9 for all the utilized datasets except for the Tiny ImageNet where it is set to 0.99. 

\subsection{Experimental Results}

	\begin{table*}
	\centering
	 \caption{Test Accuracy for all the utilized datasets}
 	    \label{batch64}
	  \begin{tabular}{|c|c|c|c|c|}
	      \hline
	 Method  &  Cifar-10 &   SVHN &  Fashion MNIST &  Tiny ImageNet\\   \hline
w/o OSAKD & 64.734\% $\pm$ 0.654\% & 88.706\% $\pm$ 0.306\% & 91.214\% $\pm$ 0.141\% & 31.050\% $\pm$ 1.550\%    \\ 
OSAKD-8NN & \bf{66.734\% $\pm$ 0.086}\% & \bf{89.567\% $\pm$ 0.436\%} &  \bf{91.49\% $\pm$ 0.006\%} & 31.282\% $\pm$ 0.506\%    \\ 
OSAKD-16NN & 65.97\% $\pm$ 0.733\% & 89.207\% $\pm$ 0.39\% & 91.418\% $\pm$ 0.124\% & \bf{31.916\% $\pm$ 0.982\%}   \\ \hline
 	  \end{tabular}
	    \end{table*}

The evaluation results are presented in Table \ref{batch64}. We use two different values of nearest neighbors (NN), that is 8NN and 16NN (abbreviated as OSAKD-8NN and OSAKD-16NN respectively), and we compare their performance with the baseline model, without applying knowledge distillation (abbreviated as w/o OSAKD). Best results are printed in bold.

First, it can be observed that the proposed method improves the baseline performance, for various network architectures, varying from lightweight ones (for example the model used in the three first datasets) to heavyweight ones (i.e. ResNet-50 in the Tiny ImageNet dataset), and also for various problems, varying from binary to multi-class problems.  
It can also been observed that even slightly better performance for the 10-class problems is also achieved considering 8NN. However, studying the marginal case -based on the number of classes (i.e. 200 classes)- of Tiny ImageNet, we can draw the conclusion that in problems consisting of considerably more classes, computing the soft labels using more nearest neighbors allows for producing more reliable soft labels that convey more useful information about the similarities of the samples with the classes.

\begin{figure*}
  \begin{subfigure}[b]{0.245\textwidth}
\includegraphics[width=\textwidth]{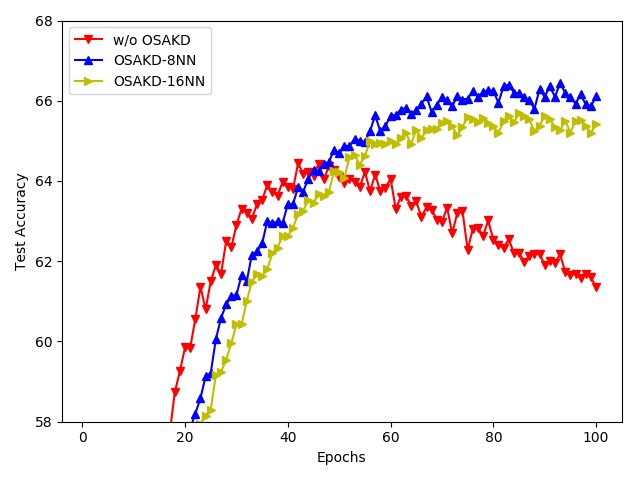}
\caption{Cifar-10}
  \label{sf1}
    \end{subfigure}
  \begin{subfigure}[b]{0.245\textwidth}
\includegraphics[width=\textwidth]{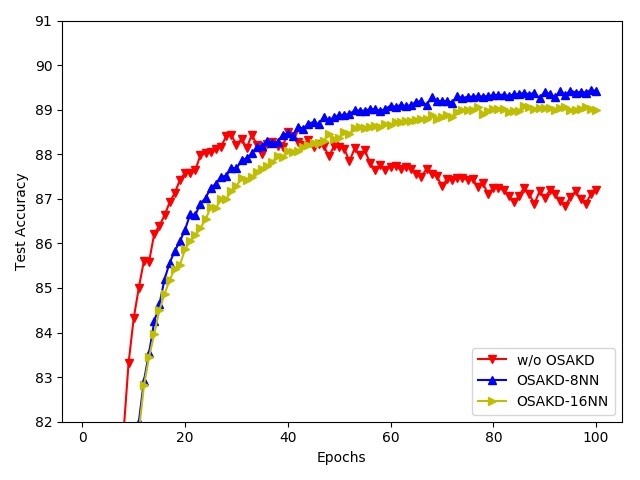}
\caption{SVHN}
  \label{sf2}
   \end{subfigure}
 \begin{subfigure}[b]{0.245\textwidth}
\includegraphics[width=\textwidth]{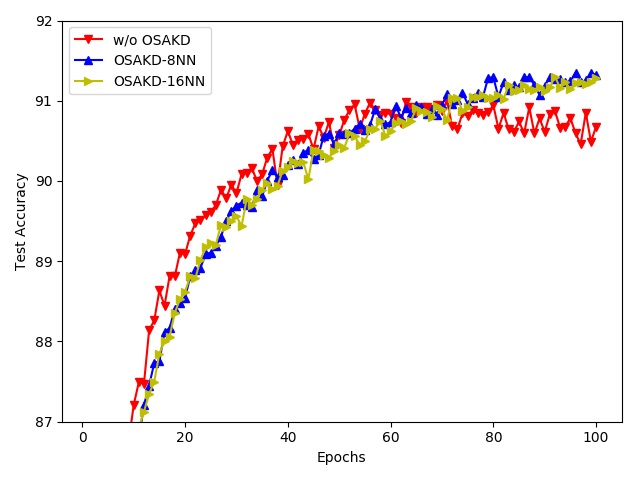}
\caption{Fashion MNIST}
  \label{sf3}
\end{subfigure}
   \begin{subfigure}[b]{0.245\textwidth}
\includegraphics[width=\textwidth]{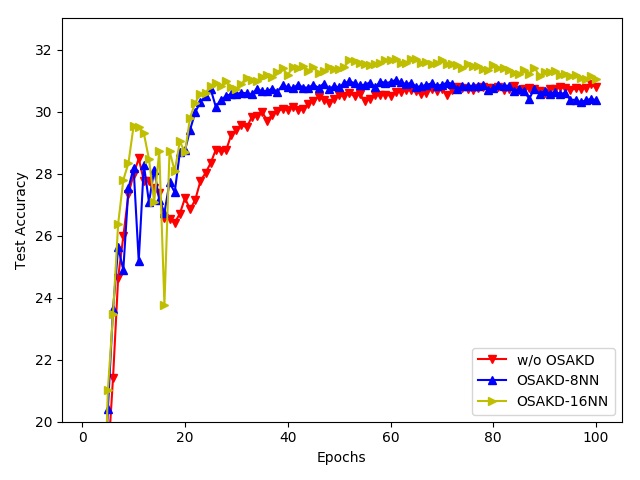}
\caption{Tiny ImageNet}
  \label{sf4}
  \end{subfigure}
  \caption{Test accuracy throughout the training epochs.}
  \end{figure*}

Figs \ref{sf1}-\ref{sf4} illustrate the comparisons of the mean test accuracy over the epochs of training of the proposed method considering 8NN and 16NN for all the utilized datasets. The enhanced performance of the OSAKD method is illustrated in the Figures. Furthermore, the generalization ability of the proposed online distillation method is clearly depicted. In addition, it is demonstrated that the proposed method provides in general more stable performance, as compared to the baseline of training without distillation. Another observation that can be made from the Figures, is that as the training process progresses, more meaningful soft labels are produced, since as we have already stated the soft labels are driven by the supervised loss, and we expect to produce more reliable soft labels throughout the training procedure.

Furthermore, we compare the proposed methods with state-of-the-art online distillation methods. More specifically, we utilize a common architecture, that is ResNet-32 \cite{DBLP:journals/corr/HeZRS15}, we apply the proposed online distillation method on Cifar-10 dataset, and compare the performance with competitive online distillation methods, \cite{NIPS2018_7980,kim2019feature}. For fair comparisons we follow the same training setup as in \cite{NIPS2018_7980,kim2019feature}. That is, we use the SGD with Nesterov momentum and set the momentum to 0.9. The initial learning rate is set to 0.1 and drops to 0.01 at 50\% training and to 0.001 at 75\%. The network is trained for 300 epochs. 

We compare the proposed method with ONE \cite{NIPS2018_7980} and FFL \cite{kim2019feature}. It should be noted that since the proposed method is a unique branch method that does not use multiple branches of the network, we use only two sub-networks in all the competitive approaches, similar to \cite{kim2019feature}, for as much as possible fair comparisons. Thus, we compare the OSAKD method with ONE distillation method, considering the average performance of the two branches, and correspondingly with the FFL-S distillation method considering the average performance of the two sub-networks. It should also be noted that the number of parameters in both FFL-S and ONE cases in the test phase is identical to the OSAKD case, since the additional branches in both cases as well as the fusion module in FFL-S are removed during the test phase. Evaluation results are presented in Table \ref{rescom}. As it is demonstrated, the proposed online distillation method is superior over competitive online distillation methods.

\begin{table}[!ht]
	   \caption{Comparisons against online distillation methods on Cifar-10 utilizing the ResNet-32 architecture.}
 	   \label{rescom}
	    \centering
	  \begin{tabular}{|c|c|c|}
	   \hline
	 Method  &  Test Accuracy\\   \hline
ResNet-32 & 93.07\% $\pm$ 0.17\% \\
ONE \cite{NIPS2018_7980} & 93.76\%$\pm$ 0.12\%   \\
FFL-S \cite{kim2019feature} & 93.81\% $\pm$ 0.12\%  \\ 
\bf{OSAKD} & \textbf{93.93\% $\pm$ 0.09\%}  \\ 
\hline
 	  \end{tabular}
	    \end{table}

Finally, we evaluate the complexity of the proposed method using the sum of floating point operations (FLOPs) in one forward pass on a fixed input size. We use the ResNet-32 model on the Cifar-10 dataset. In order to highlight the effectiveness of the proposed method we compare the complexity with the most famous offline KD method, \cite{hinton2015distilling}. In this case, we use as teacher the stronger ResNet-110 model. 
The OSAKD method requires 0.07 GFLOPs, while the offline KD requires 0.33 GFLOPs, rendering the proposed online distillation method as significantly more efficient as compared to the conventional offline methodology. We should finally note that competitive online distillation methods that utilize multiple branches or copies of a given network, require at least two times more FLOPs than the proposed one. That is, the proposed online distillation method is also more efficient as compared to competitive online methods.

%
%
\section{Conclusions}\label{s5}
In this paper, we proposed a novel single-stage online self-distillation approach, namely Online Self-Acquired Knowledge Distillation (OSAKD). The proposed method utilizes the k-nn non-parametric density estimation in order to estimate the unknown density distributions of the data samples in the output feature space. In this way, we are able to directly estimate the posterior class probabilities, revealing explicitly the similarities of the data samples with each class. This approach, allows us for deriving additional knowledge directly from the data, without affecting the model architecture by adding multiple branches or employing multiple models, and at the same time in a single stage training pipeline. The experimental evaluation indicates the effectiveness of the proposed method to improve the performance of any model, regardless its complexity.

\section*{Acknowledgment}
This research has been partially financially supported by General Secretariat for Research and Technology (GSRT) and the Hellenic Foundation for Research and Innovation (HFRI) (Scholarship Code: 2826.) and the European Union's Horizon 2020 research and innovation programme under grant agreement No 871449 (OpenDR). This publication reflects the authors' views only. The European Commission is not responsible for any use that may be made of the information it contains.

\bibliographystyle{IEEETran}
\bibliography{icme2021}

\end{document}